\documentclass{article}
\usepackage{spconf,amsmath}
\usepackage{subfig,balance}
\usepackage[pdftex]{graphicx}
\usepackage{enumitem}
\setlist{nosep, leftmargin=14pt}
\usepackage{tabularray}
\usepackage{mwe} 
\usepackage{booktabs}
\usepackage{amssymb}
\usepackage{pifont}
\usepackage{algorithm}
\usepackage{algorithmic}
\newcommand{\xmark}{\ding{55}}%
\usepackage{multirow}

\newcommand{\mmL}{\mathcal{L}}

\title{Federated Alternate Training (FAT): Leveraging Unannotated Data Silos in Federated Segmentation for Medical Imaging}
\name{Anonymous Authors}
\address{Paper under Double Blind review}
\name{Erum Mushtaq$^{\star }$ \qquad Yavuz Faruk Bakman$^{\star }$ \qquad Jie Ding$^{\dagger}$ \qquad Salman Avestimehr$^{\star }$}

\address{$^{\star}$ University of Southern California,
    $^{\dagger}$University of Minnesota}
\begin{document}
%
\maketitle
\begin{abstract}
Federated Learning (FL) aims to train a machine learning (ML) model in a distributed fashion to strengthen data privacy with limited data migration costs. It is a distributed learning framework naturally suitable for privacy-sensitive medical imaging datasets. However, most current FL-based medical imaging works assume silos have ground truth labels for training. In practice, label acquisition in the medical field is challenging as it often requires extensive labor and time costs. To address this challenge and leverage the unannotated data silos to improve modeling, we propose an alternate training-based framework, Federated Alternate Training (FAT), that alters training between annotated data silos and unannotated data silos. Annotated data silos exploit annotations to learn a reasonable global segmentation model. Meanwhile, unannotated data silos use the global segmentation model as a target model to generate pseudo labels for self-supervised learning. We evaluate the performance of the proposed framework on two naturally partitioned Federated datasets, KiTS19 and FeTS2021, and show its promising performance.

\end{abstract}

\section{Introduction}
\label{sec:intro}

In recent years, Federated Learning (FL) has been widely explored for medical applications \cite{terrail2022flamby}. However, most current works focus on supervised federated learning where all silos have pixel-wise annotations available. In practical scenarios, pixel-level label acquisition for massive medical imaging datasets requires a radiologist expert and therefore, can be time-consuming and expensive, so not all silos can afford it. Examples are silos from rural regions with limited expert resources. It has motivated us to study the research question: \textit{How can a server leverage unannotated data silos, that have no labeled data, along with a few labeled data silos in a realistic non-independent and identical (non-IID) data distribution based FL regime to improve the global model performance. Further, we focus on a more realistic scenario where the number of the unannotated data silos can be larger than the annotated data silos.}

Recently, the work of \cite{yang2021federated} studied this research problem and proposed a threshold-based self-supervised learning method to leverage unannotated data silos to segment COVID-19-affected regions. This work considered two data silos (one annotated and one unannotated).
The work of \cite{wen2022federated} used the model bank approach to extract pseudo labels from all supervised silos' models at unannotated data silos. Given the large model sizes for the 3D medical datasets, the computation of pseudo labels using several models at unannotated silos can be computationally infeasible. Another related work \cite{semiFL} studied semi-supervised federated learning in a different setting where a server has labeled data and silos have unlabeled data.

To leverage unannotated data silos, we propose a new Federated Learning framework, Federated Alternate Training (FAT), to leverage unannotated data silos. We show that a straightforward application of the centralized semi-supervised works in FL may not yield optimal results. Also, alternate training of annotated data silos and unannotated data silos is more efficient than the standard FedAvg training \cite{mcmahan2017communication} of all silos in terms of aggregation cost per round. Finally, we compare our method with the state-of-the-art method \cite{yang2021federated} and show significant improvements over it.

\section{Proposed method}
\label{sec:proposedmethod}
\begin{figure*}[htb]
\begin{minipage}{1.0\linewidth}
  \centering
  \centerline{\includegraphics[width=16cm, height=8cm]
  {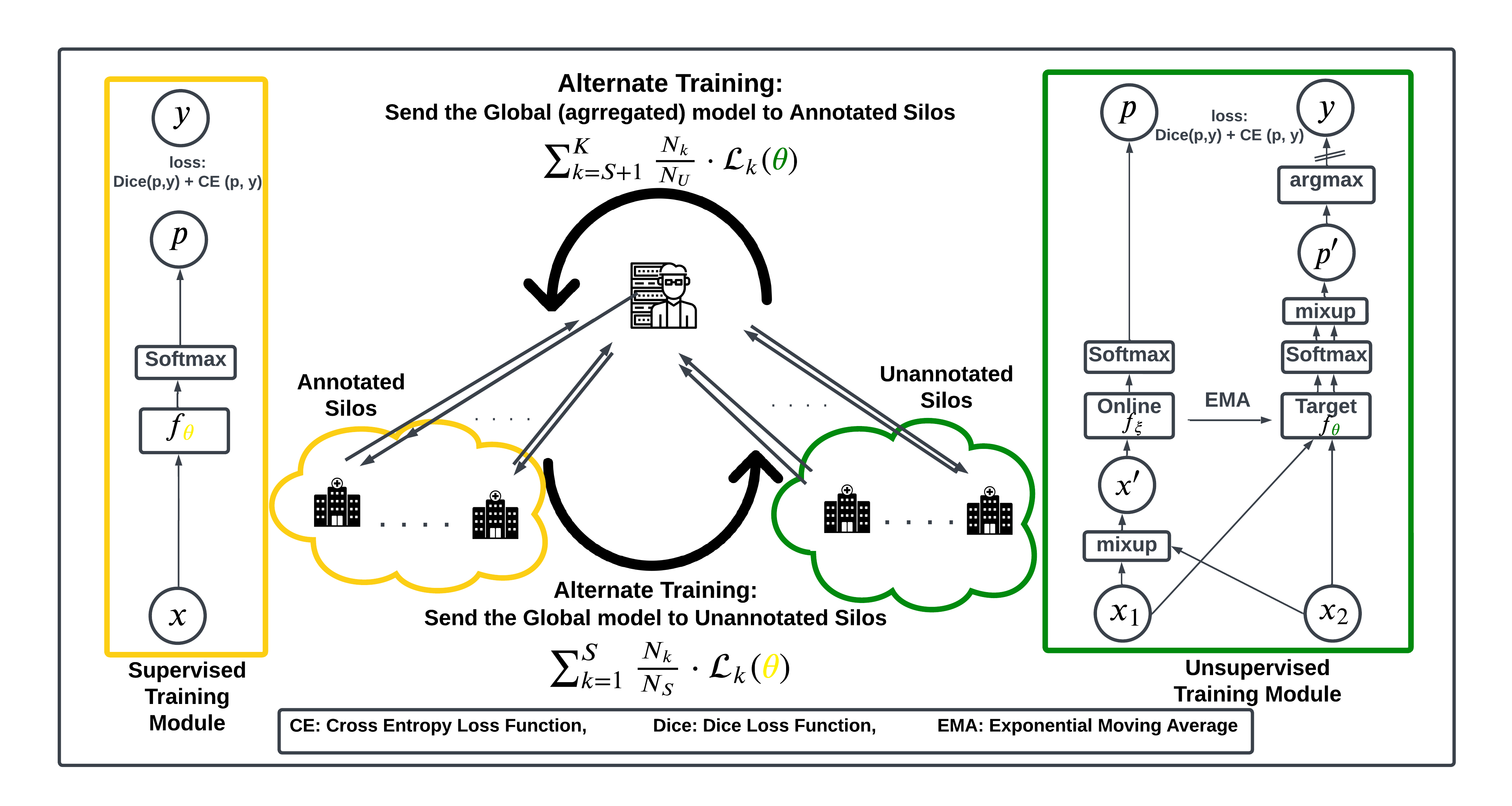}}
\end{minipage}

\caption{The proposed Federated Alternate Training (FAT) framework where we alternate training between Annotated Data Silos and Unannotated Data Silos. The Annotated Data Silos follow a supervised training module where they have ground truth labels available. The Unannotated Data Silos follow a bootstrapping-based self-supervised training module where the target model generates pseudo labels, y, for the self-supervised learning and uses exponential moving average (EMA) for the model updates.}
\label{fig:sysmodel}
\end{figure*}
\begin{algorithm}[!h]
    \caption{FAT Algorithm. }
    \begin{small}
        \begin{algorithmic}[1]
            \STATE \textbf{Initialization:} 
            ${\theta}_0$: Pretrained model weights; $s = \{1,2,...,S\}$; $u=\{S+1,S+2,...,K\}$; $E$: number of local epochs; $A$: number of rounds for supervised silos training before alternating; $\textrm{DL}$: Soft Dice loss function; $\textrm{CE}$: Cross-entropy loss function; $\tau$: weight decay; $N_i$: number of samples at client i.
            \STATE \textbf{Server runs:}
            \begin{ALC@g}
            \FOR{each round $t = 0, 1, 2, ..., T-1$}
            \IF[Supervised Round]{$(t \bmod{2A}) < A$ }  
                \FOR{each supervised client $s$ \textbf{in parallel}}
                    \STATE $\theta_{t+1}^{s}  \leftarrow \text{SupervisedTraining}(s, \theta_t)$
                \ENDFOR
                \STATE $\theta_{t+1} \leftarrow \sum_{s=1}^{S} \frac{N_{s}}{N_{S}} \theta_{t+1}^{s}$, $N_{S} = \sum_{s=1}^{S} N_{s}$
            \ELSE[Unsupervised Round]
                \FOR{each unsupervised client $u$ \textbf{in parallel}}
                    \STATE $\theta_{t+1}^{u}  \leftarrow \text{UnsupervisedTraining}(u, \theta_t)$
                \ENDFOR
                \STATE $\theta_{t+1} \leftarrow \sum_{u=S+1}^{K} \frac{N_{u}}{N_U} \theta_{t+1}^{u}$, $N_{U} = \sum_{u=S+1}^{K} N_{u}$      
            \ENDIF
            \ENDFOR
            \end{ALC@g}
            \STATE \textbf{SupervisedTraining}($s$, $\theta$): // \textit{Supervised client $s$}
            \begin{ALC@g}
            \FOR{$e$ in epoch E}
            \FOR{minibatch $x$ in training data}
            \STATE $\mmL_{\mathrm{tr}}(\theta) = \textrm{DL}(p, y) + \textrm{CE}(p, y)$, $p = f_{\theta}(x)$
            \STATE Update $\theta = \theta - \gamma_\theta \nabla_\theta\mmL_{\mathrm{tr}}(\theta)$
            \ENDFOR
            \ENDFOR \quad return $\theta$ to server
            \end{ALC@g}
            \STATE \textbf{UnsupervisedTraining}($u$, $\theta$): // \textit{Unsupervised client $u$}
            \begin{ALC@g}
            \STATE $\xi \leftarrow \theta$
            \FOR{$e$ in epoch E}
            \FOR{sample two batches ($x_1$, $x_2$) in training data}
            \STATE $p = f_{\xi}(x')$, $x' = \lambda x_1 + (1 - \lambda) x_2$
            \STATE $p_{1} = f_{\theta}(x_1)$, $\quad$ $p_{2} = f_{\theta}(x_2)$
            \STATE $y = argmax(p')$, $p' = \lambda p_1 + (1 - \lambda) p_2$
            \STATE $\mmL_{\mathrm{tr}}(\xi) = \textrm{DL}(p, y) + \textrm{CE}(p, y)$
            
            \STATE Update $\xi = \xi - \gamma_\xi 
            \nabla_\xi\mmL_{\mathrm{tr}}(\xi)$
            \STATE Update $\theta = \tau \theta + (1 - \tau ) \xi$
            \ENDFOR
            \ENDFOR \quad return $\theta$ to server
            \end{ALC@g}
        \end{algorithmic}
    \end{small}
\label{alg:fssl}
\end{algorithm}

Federated Optimization focuses on a distributed optimization task where K nodes collaborate with each other to learn a global model with parameters $\theta$ as shown below,
\begin{equation}
    \min_{\theta} G(\mathcal{L}_{1}(\theta; X_{1}, Y_{1}), ..., \mathcal{L}_{K}(\theta; X_{K}, Y_{K})),
\end{equation}
where $\mathcal{L}_{i}(\theta; X_{i}, Y_{i})$ represents node i's local loss function, $X_{i}$ denotes the training data and $Y_{i}$ represents the labels at node $i$. $G(.)$ can be any function, for example, $G(.)$ aggregates the local objectives $(\sum_{k=1}^{K} \frac{N_{k}}{N} \cdot \mathcal{L}_k(\theta; X_{k}, Y_{k}))$ in Federated Averaging algorithm \cite{mcmahan2017communication}, where $N_{k}$ is the total number of training data samples at node $k$ and $\sum_{k=1}^{K} N_{k} = N$.
\subsection{Problem Formulation}
In a typical federated averaging setting, each node consists of annotated data ($X_{i}$, $Y_{i}$). However, it is very unlikely that all nodes have labeled data. Some nodes may not have any labels at all. In such a setting, the optimization problem becomes, 
\begin{equation}
\begin{split}
    &\min_{\theta \in \mathbb{R}^d} G(\mathcal{L}_{1}(\theta; X_{1}, Y_{1}),\mathcal{L}_{2}(\theta; X_{2}, Y_{2}),..., \mathcal{L}_{S}(\theta; X_{S}, Y_{S}), \\ &\mathcal{L}_{S+1}(\theta; X_{S+1}),\mathcal{L}_{S+2}(\theta; X_{S+2}),..., \mathcal{L}_{K}(\theta; X_{K})),
    \end{split}
    \label{objective}
\end{equation}
where $S$ signifies the number of nodes. Nodes in the supervised silo $\{1,2,.., S\}$ contain annotated data i.e, both $X$ and $Y$. The rest of the nodes, $\{S+1, S+2,.., K\}$, are unsupervised and therefore contain unlabelled data i.e, only $X$. Hence, the objective is to learn a global model such that learning from the unsupervised nodes $\{S+1,S+2, .., K\}$ nodes along with supervised nodes $\{1,2,.., S\}$ increases the global model performance as compared to the global model learned from all the supervised nodes alone. 


\subsection{Federated Alternate Training (FAT)}
We proposed alternate training between the supervised and unsupervised silos to solve the objective in eq. (\ref{objective}). In the first round, we initialize our global model with the models pre-trained on other medical datasets. We send this model to the supervised silos, which will fine-tune the global model using their labeled data. The global objective $G(.)$ aggregates the model weights obtained from the supervised silos, $\sum_{k=1}^{S} \frac{N_{k}}{\sum_{i=1}^{S} N_{i}} \cdot \mathcal{L}_k(\theta)$ and send it to unsupervised silos where it is used to obtain pseudo-labels for learning. After this round, the global objective $G(.)$ aggregates the model weights sent by the unannotated data silos, $\sum_{k=S+1}^{K} \frac{N_{k}}{\sum_{i=S+1}^{K} N_{i}} \cdot \mathcal{L}_k(\theta)$. Hence, the objective $G(.)$ alternates between aggregating the supervised silos model weights for a few rounds and the unsupervised silos model weights for the next few rounds. Next, we explain how we obtain pseudo labels at the unsupervised silos.

\subsection{Bootstrapping}
We perform self-supervised learning in unsupervised silos.
During self-supervised training, we aim to learn from the global model without forgetting what the global model learned from the supervised silos. To that end, we bootstrap the learned labels. Instead of maintaining one neural architecture, as is used in the previous work \cite{yang2021federated}, we maintain two models referred to as the online model with parameters $\xi$ and the target model with parameters $\theta$.
Unsupervised silos initialize both models with the global model at the start of each round. 
For self-supervised training, we use the mixup approach \cite{basak2022embarrassingly} to augment the input data and feed the perturbed version $x' = \lambda x_1 + (1 - \lambda) x_2$ of two randomly selected input data points $x_1$ and $x_2$, where $\lambda \in (0, 1)$ and is a hyperparameter. We feed $x'$ to the online network, $f_{\xi}$. The online model outputs each class's prediction probabilities $p$ for the perturbed input $x'$. In parallel, we feed the unperturbed data points to $x_1$ and $x_2$ to the target model $f_{\theta}$ and perturbed their corresponding prediction probabilities $p_{1}$ and $p_{2}$ via mixup logic $ p' = \lambda p_1 + (1 - \lambda) p_2$.
The pseudo label $y$ is obtained by applying the argmax operation on the perturbed output $p'$. These pseudo labels are used to train the online model via Dice loss and Cross Entropy Loss between the pseudo label $y$ and $p$. After each training step of the online model, the target model is updated by the exponential moving average $\theta = \tau \theta + (1 - \tau) \xi$,
where $\tau \in (0, 1)$ is a decay rate of the target model. At the supervised silos, we do not need pseudo labels. Thus, we train only one model with parameters $\theta$ and use Dice loss and Cross Entropy loss between the ground truth label $y$ and the predicted probabilities $p$. The overall framework is shown in Figure \ref{fig:sysmodel} and Algorithm \ref{alg:fssl}.

\begin{table*}[htb]
\centering
    \caption{Comparison of Different Learning Methods: KiTS19 and FETS2021 Dataset}
\resizebox{0.9\textwidth}{!}{
    \centering
    \begin{tabular}{cccccc}
    \toprule
     \multirow{2}{*}{\textbf{Method}} & \multicolumn{2}{c}{\textbf{KiTS19}} &  \multicolumn{3}{c}{\textbf{FeTS2021}} \\
     & \textbf{Kidney Dice Score} & \textbf{Tumor Dice Score} & \textbf{WT Dice Score} & \textbf{TC Dice Score} & \textbf{ET Dice score}\\
     \midrule
        {Fully Supervised - Centralized (U+S)}  & 0.949 & 0.750 & 0.929& 0.798 & 0.66\\ 
        \midrule
       {Fully Supervised - FL (U+S)}  & 0.940& 0.717 & 0.913 & 0.781 &  0.644\\
       {Fully Supervised - FL + PreTraining (U+S)}  & 0.951& 0.781 & 0.919 & 0.795 & 0.661 \\
       {Fully Supervised - FL + PreTraining (S)}  & 0.929& 0.553 & 0.910 & 0.777 & 0.614  \\
        \midrule 
        Semi-supervised - Centralized (U+S) & 0.937 & 0.712  & 0.927 & 0.793 & 0.655\\ 
       Semi-supervised - FL  [ \cite{yang2021federated}] - (U+S) & 0.943 & 0.615 & 0.912 & 0.773 & 0.635  \\
       Semi-supervised - FAT - [Ours] - (U+S ) & \textbf{0.951} & \textbf{0.730} & \textbf{0.913} & \textbf{0.780} & \textbf{0.660}  \\
      \bottomrule
    \end{tabular}}
    \label{tab:accuracy comparison3}
\vspace{-0.3cm}
\end{table*}

\subsection{Datasets and Experimental Setup}
\begin{figure}
    \centering
    \subfloat[KiTS19 Data Distribution across Silos \label{fig:kits19dataset}]{\includegraphics[width=6cm, height = 4.2cm]{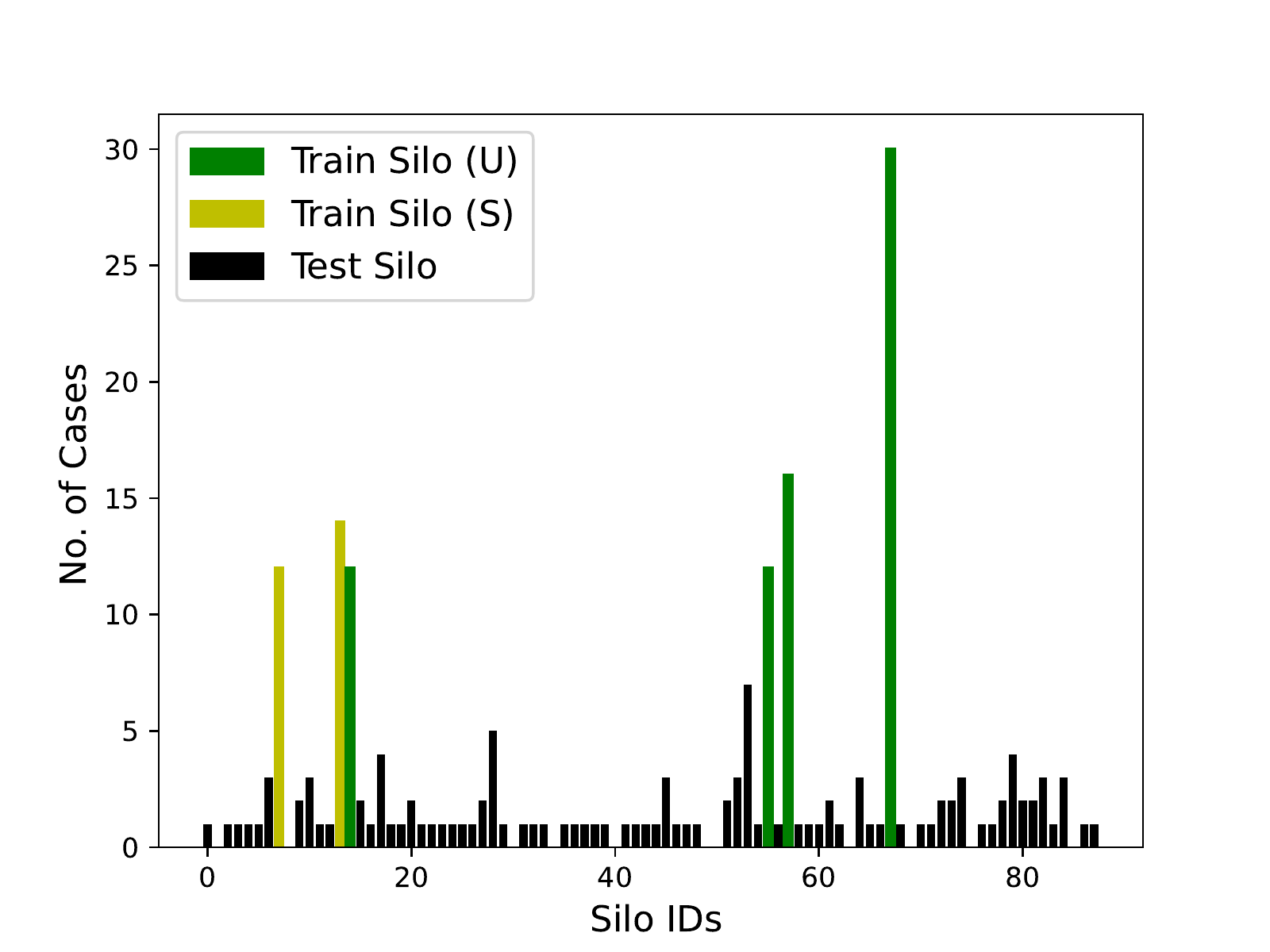}}%
    \qquad
    \subfloat[FETS2021 Data Distribution across Silos \label{fig:fets2021}]{\includegraphics[width=6cm, height = 4.2cm]{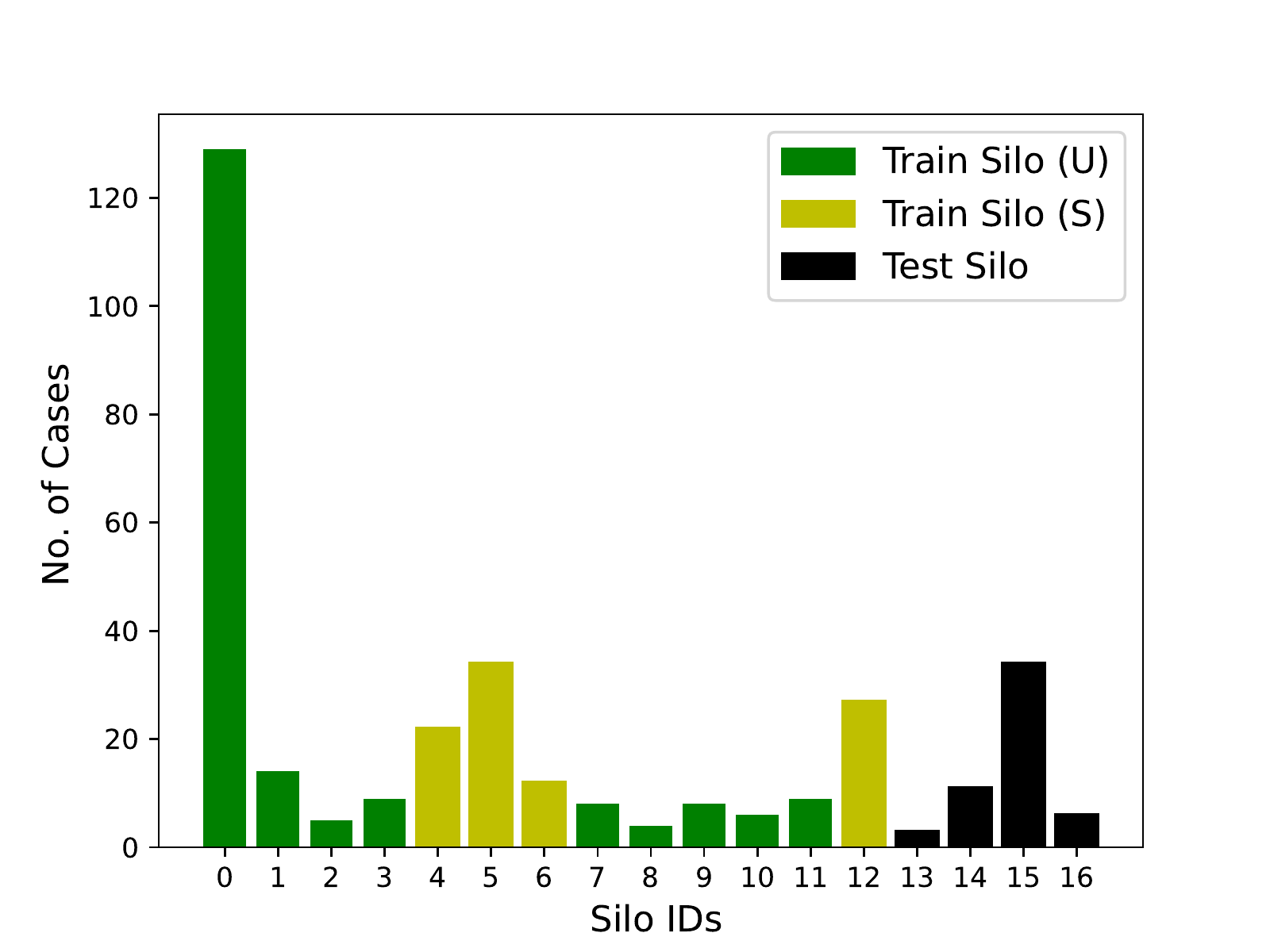}}%
    \caption{Data distribution in terms of Supervised (S) and Unsupervised (U) Train Silos, and Test Silos. 
}%
    \label{fig:datadistribution}%
\end{figure}

We evaluate the performance of the proposed framework over two
public, naturally partitioned medical datasets, KiTS19 \cite{heller2019kits19, heller2020state} and FeTS2021 \cite{pati2021federated, reina2021openfl, bakas2017advancing}. We follow \cite{terrail2022flamby} to obtain the federated version of KiTS19, which give us 6 silos as training silos and the rest of the silos are used as test silos. Since we focus on global semi-supervision learning, we further split the train silos into two supervised silos (S) and four unsupervised silos (U) in Figure \ref{fig:kits19dataset}. For the FeTS2021 dataset, we use 13 silos as training silos and 4 silos as test silos. We further split the train silos into four supervised silos and nine unsupervised silos in Figure \ref{fig:fets2021}. The task for FETS2021 is to segment the whole tumor (WT), enhancing tumor (ET) and tumor core (TC), whereas, the task for KiTS19 consists of segmenting the Kidney and Tumor in abdomen CT scans. We use the DICE score as our evaluation metric.
    
For preprocessing and training, we use the nnUNet pipeline and model architecture \cite{isensee2021nnu}. For model initialization, we use a nnUNet pretrained on LiTS \cite{bilic2019liver} and ACDC \cite{bernard2018deep} dataset for KiTS19 and FETS2021, respectively. For all FL experiments, we use 3000 rounds with 5 local epochs. For FAT, we alternate training after every 5 rounds. To evaluate the SoTA method \cite{yang2021federated}, we followed their approach and trained model first at the supervised silos for 500 rounds. For the remaining 2500 rounds, the unsupervised silos also participate. To keep comparison fair, we used the pretrained model based initialization for both SOTA and our method FAT at round 0. Further, we used random-intensity shift data augmentation with a level of 0.9, as given in their work.

\section{Results}
\label{sec:results}

 \begin{table}[h!]
\centering
    \caption{Ablation Study on the different components of the Proposed Scheme: KiTS19 Dataset}
\resizebox{0.5\textwidth}{!}{
    \centering
    \begin{tabular}{ccccc}
    \toprule
    \textbf{Mixup}  & \textbf{Alternate Training} & \textbf{Kidney Dice Score} & \textbf{Tumor Dice Score} \\
    \cmidrule{1-4}
            {\checkmark}   & \xmark  & 0.945 & 0.712  \\
    
      
       \cmidrule{1-4}
       {\checkmark}  & \checkmark & \textbf{0.955} & \textbf{0.730} 
       \\
      \bottomrule
    \end{tabular}}
    \label{tab:accuracy comparison3}
\vspace{-0.3cm}
\end{table}

    
      

    \begin{figure}[htb]
    \centering
    \subfloat[KiTS19: Tumor Dice Score   \hspace{1.0cm}    (b)  FeTS2021: ET Dice Score]{\includegraphics[width=8.5cm, height = 4.5cm]{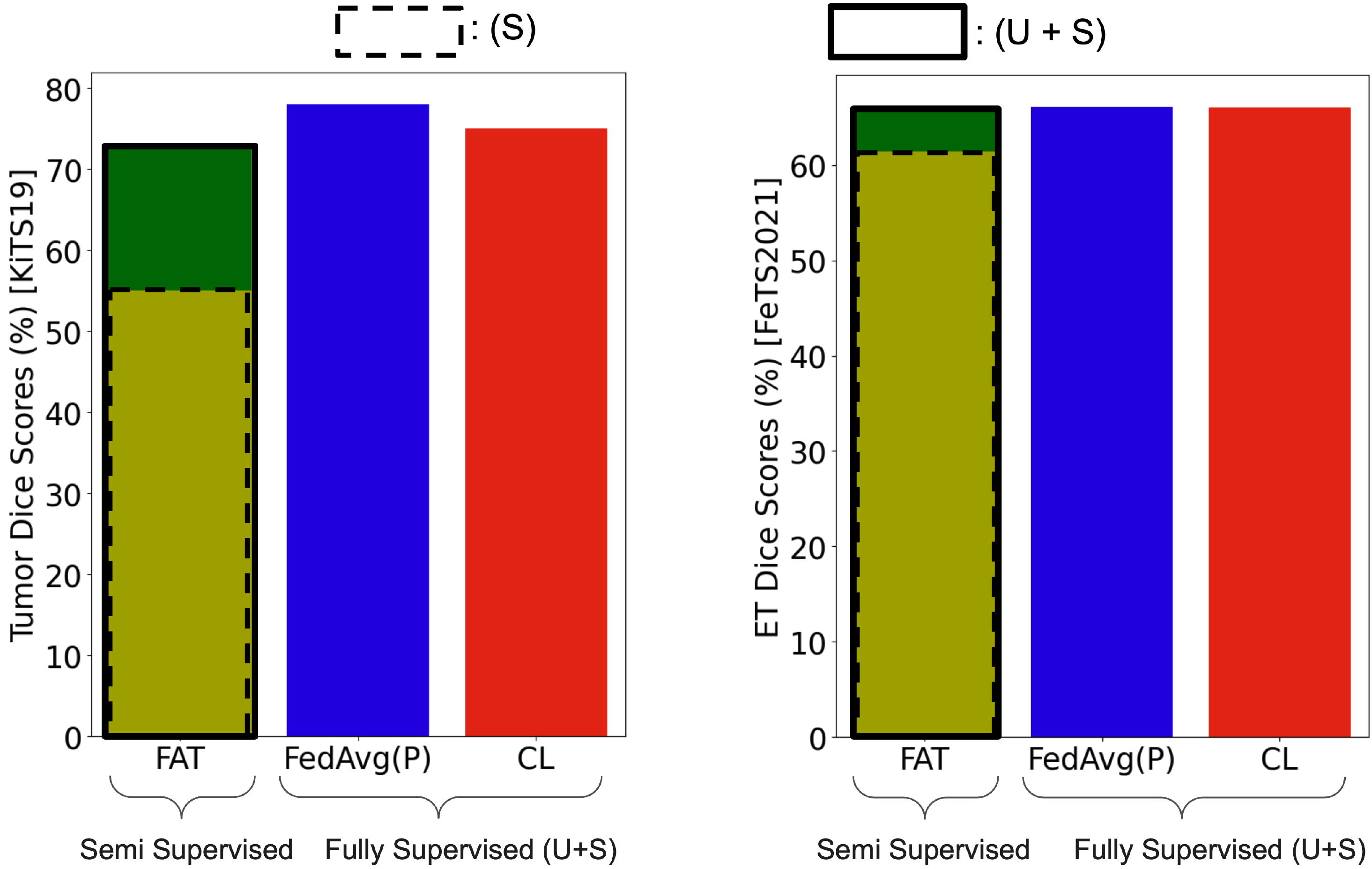}}%
    \qquad
    \caption{Dice Score comparison of the proposed framework FAT with the Fully Supervised Centralized Learning (CL) and FedAvg (P: PreTraining) benchmarks. }
    \label{fig:ccomp}
    \label{fig:comparisonfig}%
\end{figure}

\subsection{Experimental Results}
In the KiTS19 dataset, we achieve 75\% Tumor and 95\% Kidney Dice scores with a fully supervised centralized learning (CL) setting. 
For FL without pretraining, we observe a 3.3\% and 0.9\% Tumor and Kidney Dice Score drop compared to CL. However, with pretrained model initialization, we achieve promising results for FL, 78\% Tumor, and 95\% Kidney Dice score. We found that with pretraining- based FL, we can even outperform centralized training. Therefore for all our semi-supervised learning experiments, we initialize our model with a pretrained model.

For global semi-supervised learning experiments, we use the proposed framework FAT to save data annotations cost at some silos (4 out of 6) and achieve a Dice score of 95.5\% and 73\% on the Kidney and Tumor, respectively. These results can be appreciated by comparing it to not only the best method in the literature \cite{yang2021federated}, but also to Fully Supervised FL with only supervised silos (S). Our method outperforms the state-of-the-art by 10.2\% Dice score margin in Tumor and 0.8\% Dice score margin in Kidney Dice scores. Our results also demonstrate the usefulness of unsupervised silos as they can add 18\% Dice score improvement in Tumor and 3\% Dice score improvement in Kidney segmentation.


In Table \ref{tab:accuracy comparison3}, we perform an ablation study on the KiTS19 dataset where we compare the benefit of alternate training component. Without alternate training, we achieve 71.2\% Tumor Dice. However, with alternate training, we achieve 73\% Tumor Dice Score. Note that there are three main differences between SOTA method and our proposed method. We use mixup logic as a data augmentation scheme. We also use bootstrapping, student-teacher, framework. Further, we exploit alternate training. Even if we make the other two components same, that is, we use mixup with student-teacher framework but without alternate training, we achieve 71.2\% Tumor Dice score which is 1.8\% lower than the FAT performance. This highlights the importance of the proposed algorithm.

In the FeTS2021 dataset, we achieve 93\% WT, 80\% TC, and 66\% ET Dice scores with a fully supervised centralized learning setting. For FL with pretraining, we achieve 92\% WT, 80\% TC, and 66\% ET Dice score, which shows comparable performance with the CL benchmark. Further, to save data annotations cost at 9 silos of 13, we achieve a Dice score of 91\% WT, 78\% TC, 66\% ET Dice Scores with our proposed algorithm FAT. Our method outperforms the state-of-the-art by 0.7\% TC Dice score margin and 2.5\% Dice score margin in ET Dice scores. Our results also demonstrate the usefulness of unsupervised silos as they can add 5\% Dice score improvement in ET segmentation as also shown in Fig. \ref{fig:ccomp}.

\section{Conclusion}
In this work, we proposed a novel federating learning framework, FAT, for medical segmentation tasks. FAT exploits both the supervised and unsupervised silos by alternating training between them. As a result, it can leverage unsupervised silos to enhance the global model performance and outperform the state-of-the-art method in KiTS19 and FeTS2021 datasets. 
\label{sec:acknowledgments}
\section{Acknowledgements}
This material is based upon work supported by Defense Advanced Research Projects Agency (DARPA) under Contract No. FA8750-19-2-1005. The views, opinions, and/or findings expressed are those of the author(s) and should not be interpreted as representing the official views or policies of the Department of Defense or the U.S. Government. This work is also supported by research gifts from Intel and Konica Minolta.
\section{COMPLIANCE WITH ETHICAL STANDARDS}
This research study was conducted retrospectively using hu- man subject data made available in open access. Ethical ap- proval was not required as confirmed by the license attached with the open-access data.

\balance
\bibliographystyle{IEEEbib}
\bibliography{strings,refs}

\end{document}